\documentclass[letterpaper]{article} % DO NOT CHANGE THIS
\usepackage{aaai23}  % DO NOT CHANGE THIS
\usepackage{times}  % DO NOT CHANGE THIS
\usepackage{helvet}  % DO NOT CHANGE THIS
\usepackage{courier}  % DO NOT CHANGE THIS
\usepackage[hyphens]{url}  % DO NOT CHANGE THIS
\usepackage{graphicx} % DO NOT CHANGE THIS
\usepackage{natbib}  % DO NOT CHANGE THIS AND DO NOT ADD ANY OPTIONS TO IT
\usepackage{caption} % DO NOT CHANGE THIS AND DO NOT ADD ANY OPTIONS TO IT
\usepackage{pdfpages}
\usepackage{makecell}
\usepackage{multirow}
\usepackage{amssymb}
\usepackage{algorithm}
\usepackage{algorithmic}
\usepackage{lineno}
\usepackage{amsmath}
\usepackage{xcolor}
\newcommand\modelname[1]{X4D-SceneFormer}
\usepackage{newfloat}
\usepackage{listings}
\DeclareCaptionStyle{ruled}{labelfont=normalfont,labelsep=colon,strut=off} % DO NOT CHANGE THIS
\floatstyle{ruled}
\newfloat{listing}{tb}{lst}{}
\floatname{listing}{Listing}
\nocopyright
\title{X4D-SceneFormer: Enhanced Scene Understanding on 4D Point Cloud Videos through Cross-modal Knowledge Transfer}
\author{
    Linglin Jing\textsuperscript{\rm 3,1}\equalcontrib, Ying Xue\textsuperscript{\rm 2}\equalcontrib, Xu Yan\textsuperscript{\rm 2$\dagger$}, Chaoda Zheng\textsuperscript{\rm 2}, Dong Wang\textsuperscript{\rm 1}, 
    Ruimao Zhang\textsuperscript{\rm 2},\\ Zhigang Wang\textsuperscript{\rm 1}, Hui Fang\textsuperscript{\rm 3},
    Bin Zhao\textsuperscript{\rm 1}, 
    Zhen Li\textsuperscript{\rm 2}\thanks{Corresponding authors: Xu Yan and Zhen Li.}\\
}
\affiliations{
    \textsuperscript{\rm 1}Shanghai Artifcial Intelligence Laboratory,\\
    \textsuperscript{\rm 2}The Chinese University of Hong Kong (Shenzhen), 
    
    \textsuperscript{\rm 3}Department of Computer Science, Loughborough University

   \texttt{\small\{jinglinglin\}@pjlab.org.cn, \{xuyan1@link., lizhen@\}cuhk.edu.cn}

}

\usepackage{bibentry}
\begin{document}

\maketitle
% \linenumbers

\begin{abstract}
% 4D Temporal action segmentation (TAS) from point cloud video aims at densely identifying point cloud frames in minutes-long videos with multiple action classes has attracted a lot of attention. Because of the irregularities in point cloud videos where points emerge inconsistently across different frames, identifying all frames in long untrimmed videos often presents boundary blurring and over-segmentation problems. To address
% the above issue, in this paper, we introduce a decomposed cross-modal distillation framework by transferring knowledge of the 2D appearance and motion modality to 3D point clouds. Specifically, instead of direct distillation, we propose a dual-branch design to separately distill 2D appearance and motion representations to assist 3D representation in training stage. Finally, several contrasive losses are introduced to solve the over-segmentation problem. Extensive experiments and ablations on 4D temporal action segmentation demonstrate the effectiveness of our approach for point cloud video modeling.

% Recent work on 4D point cloud videos has attracted
% a lot of attention. Because of the irregularities in point cloud videos where points emerge inconsistently across different frames, identifying all frames in long untrimmed videos often presents boundary blurring.

The field of 4D point cloud understanding is rapidly developing with the goal of analyzing dynamic 3D point cloud sequences.
However, it remains a challenging task due to the sparsity and lack of texture in point clouds. 
Moreover, the irregularity of point cloud poses a difficulty in aligning temporal information within video sequences.
To address these issues, we propose a novel \textbf{cross}-modal knowledge transfer framework, called \textbf{\modelname,}. 
This framework enhances \textbf{4D-Scene} understanding by transferring texture priors from RGB sequences using a \textbf{Transformer} architecture with temporal relationship mining.
Specifically, the framework is designed with a dual-branch architecture, consisting of an 4D point cloud transformer and a Gradient-aware Image Transformer (GIT).
The GIT combines visual texture and temporal correlation features to offer rich semantics and dynamics for better point cloud representation.
During training, we employ multiple knowledge transfer techniques, including temporal consistency losses and masked self-attention, to strengthen the knowledge transfer between modalities. This leads to enhanced performance during inference using single-modal 4D point cloud inputs.
Extensive experiments demonstrate the superior performance of our framework on various 4D point cloud video understanding tasks, including action recognition, action segmentation and semantic segmentation.
The results achieve \textbf{1st places}, \textit{i.e.,} \textbf{85.3\%} (+7.9\%) accuracy and \textbf{47.3\%} (+5.0\%) mIoU for 4D action segmentation and semantic segmentation, on the HOI4D challenge\footnote{\url{http://www.hoi4d.top/}.}, outperforming previous state-of-the-art by a large margin. We release the code at \url{https://github.com/jinglinglingling/X4D} 

% The field of scene understanding on 4D point cloud videos, consisting of consecutive 3D point cloud frames, is relatively new and still lacks comprehensive research compared to the extensive work on 3D point clouds. Existing methods on 4D point cloud videos often suffer from low precision due to the irregularities in point cloud videos, where points emerge inconsistently across different frames, leading to inconsistent spatial and temporal information in their network structures. Furthermore, cross-modal approaches successful in 3D face challenges when applied directly to 4D point cloud videos, as they typically consider geometry from static snapshots only. To address these challenges, our paper introduces a novel 4D cross-modal distillation framework that transfers knowledge from RGB and motion modalities (Temporal Gradient) to 3D point clouds. The dual-branch design, incorporating RGB features and texture\&motion correlation features with a temporal sliding window, enhances the temporal consistency of motion information. Additionally, several temporal alignment contrastive losses (Supervised\&Unsupervised) improve the consistency in both multi-modals and the relationship between frames within each sequence. Experimental results demonstrate the superior performance of our framework compared to previous approaches across various 4D point cloud video understanding tasks, including action segmentation and semantic segmentation.

\end{abstract}

%%%%%%%%% BODY TEXT

\begin{figure}[!t] 
\begin{center}
\includegraphics[width=0.85\linewidth]{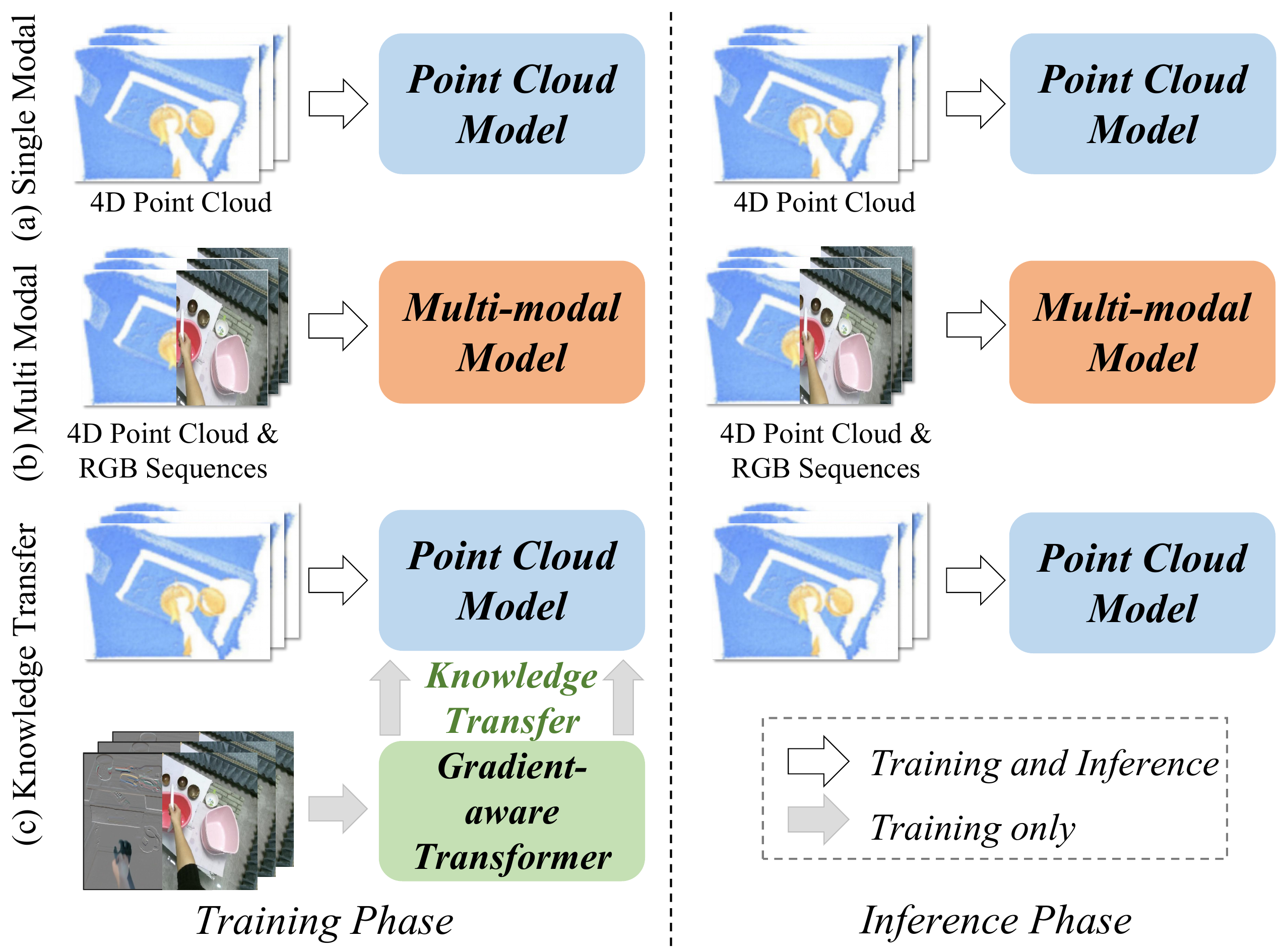}
\end{center}
\caption{\textbf{4D Cross-Modal Knowledge Transfer.} (a) Previous 4D point cloud analysis methods take point cloud only as their input. (b) Although cross-modal approaches enhance performance, they introduce extra computation overhead in both training and inference. (c) Our method takes additional 2D images during the training for 4D cross-modal knowledge transfer. During the inference, the point cloud model can be independently deployed.}
%\vspace{-3mm}
\label{fig:overview}
\end{figure}

\section{Introduction}
\label{sec:Intro}

Exploring point cloud sequences in 4D (integrating 3D space with 1D time) has garnered considerable interest in recent years~\cite{fan2021point, wen2022point, liu2022hoi4d} due to their capacity to offer a wealth of dynamic information within our 3D environment.
%
% Point cloud videos are typically captured by a 3D sensor, such as a LiDAR or RGBD camera, and consist of a sequence of 3D point clouds with a temporal dimension.
%
Compared to conventional videos, 4D point clouds deliver direct access to geometric information in 3D space, a facet particularly advantageous for real-world interactions.
%
% In contrast with 3D point clouds, 4D point cloud videos encompass temporal information, enabling the exploitation of object motion and the handling of occlusion. 
%
These attributes are pivotal for understanding 3D dynamic environments, including tasks like action recognition/segmentation~\cite{hoai2011joint}, and 4D semantic segmentation~\cite{xie2020linking}.

Previous works in 4D point cloud representation learning predominantly stem from extending existing 3D point cloud models~\cite{fan2021point, wen2022point} to 4D, which involves incorporating additional temporal learning modules that enable feature interactions across time~\cite{xiao2022learning, zhong2022no}.
%
% However, these methods encounter a limitation in capturing comprehensive semantic details due to the sparsity and lack of texture in point clouds. Yet, this capability remains crucial, particularly for tasks requiring meticulous reasoning.
However, due to the sparsity and lack of texture in point clouds, these methods are limited in capturing comprehensive semantic details. Nevertheless, such semantic information remains crucial, particularly for tasks demanding meticulous reasoning, such as 4D semantic segmentation and action segmentation.
A plausible approach to address this limitation involves integrating supplementary texture information from RGB images to enhance the 4D point cloud representation, similar to methods employed in previous cross-modal studies~\cite{cui2021deep, yan20222dpass}.
Nevertheless, as shown in Figure~\ref{fig:overview}(b), the concurrent processing of data from two modalities unavoidably introduces additional network designs and computational overhead, posing challenges in online 4D video tasks.

In this paper, we present a solution to address the aforementioned challenges through cross-modal knowledge transfer. This approach efficiently transfers color and texture-aware knowledge from 2D images to an arbitrary point-based model, while avoiding extra computational cost during the inference phase, as depicted in Figure~\ref{fig:overview}(c).
Our framework stands apart from prior cross-modal approaches that solely focus on knowledge transfer between static frame pairs~\cite{crasto2019mars}.
Notably, it places extra emphasis on ensuring motion and temporal alignment during the knowledge transfer.

The proposed framework, \textbf{\modelname{1}}, takes multi-modal data (\textit{i.e.,} 4D point cloud, RGB sequences) as input during training, and achieves superior performance using only point cloud data during inference. 
Specifically, there are two branches in our training framework, processing point cloud and RGB sequence independently.
For the point branch, we simply deploy an off-the-shelf 4D point cloud processor for the sake of simplicity.
For the image branch, we introduce a Gradient-aware Image Transformer (GIT) to learn strong image semantics. 
%
% Besides, GIT considers the temporal gradient (TG) among adjacent image frames for additional motion understanding, with contrastive losses regarding the motion and temporal alignment.
GIT takes into account the temporal gradient (TG) as an added input from adjacent image frames to enhance its comprehension of motion dynamics. Additionally, multi-level consistency losses are introduced to address both motion-related aspects and temporal alignment.
Subsequently, the semantic and motion features are integrated into a unified visual representation through cross-attention. These merged representations are then combined with the extracted point cloud representations, forming a stacked input for further processing with a cross-modal transformer.
By employing carefully-designed attention masks, the cross-modal transformer can be deployed with only point cloud inputs during inference, while still incorporating multi-modal knowledge.
In such a manner, it achieves significant improvements in effectively leveraging multi-modal information and ensuring consistent motion alignment, making it a promising solution for various 4D point cloud tasks.

In summary, the contributions of this work are:

\begin{itemize}
\item \textit{\textbf{Generality:}} We propose \modelname{1}, the first cross-modal knowledge transfer architecture for 4D point cloud understanding, where arbitrary point-based models can be easily integrated into this framework for cross-modal knowledge transfer. 

\item \textit{\textbf{Flexibility:}} We propose Gradient-aware Image Transformer (GIT) to provide temporal-aware and texture-aware features guidance. We also propose multi-level consistency metrics, employing a cross-modal transformer, to enhance knowledge transfer for the point cloud model. Notably, these techniques are only applied during training, ensuring that the point cloud model can be independently deployed during inference.

\item \textit{\textbf{Effectiveness:}} Extensive experiments on three tasks show that our method outperforms previous state-of-the-art methods by a large margin. This highlights the superiority of our approach in 4D point cloud understanding.

% Specifically, the results on the tasks of 4D xaction segmentation, 4D semantic segmentation and 4D action recognition demonstrate significant performance gains upon previous approaches, \textit{i.e.,} 85.2\% (+7.8\%) accuracy and 91.0\% (+11.4\%) edit on HOI4D action segmentation, 47.3\% (+5.0\%) mIoU on HOI4D semantic segmentation, and +3.34\% accuracy on MSR-Action3D.

\end{itemize}

\begin{figure*}[!t] 
\begin{center}
% \fbox{\rule{0pt}{2in} \rule{.9\linewidth}{0p
\includegraphics[width=1\linewidth]{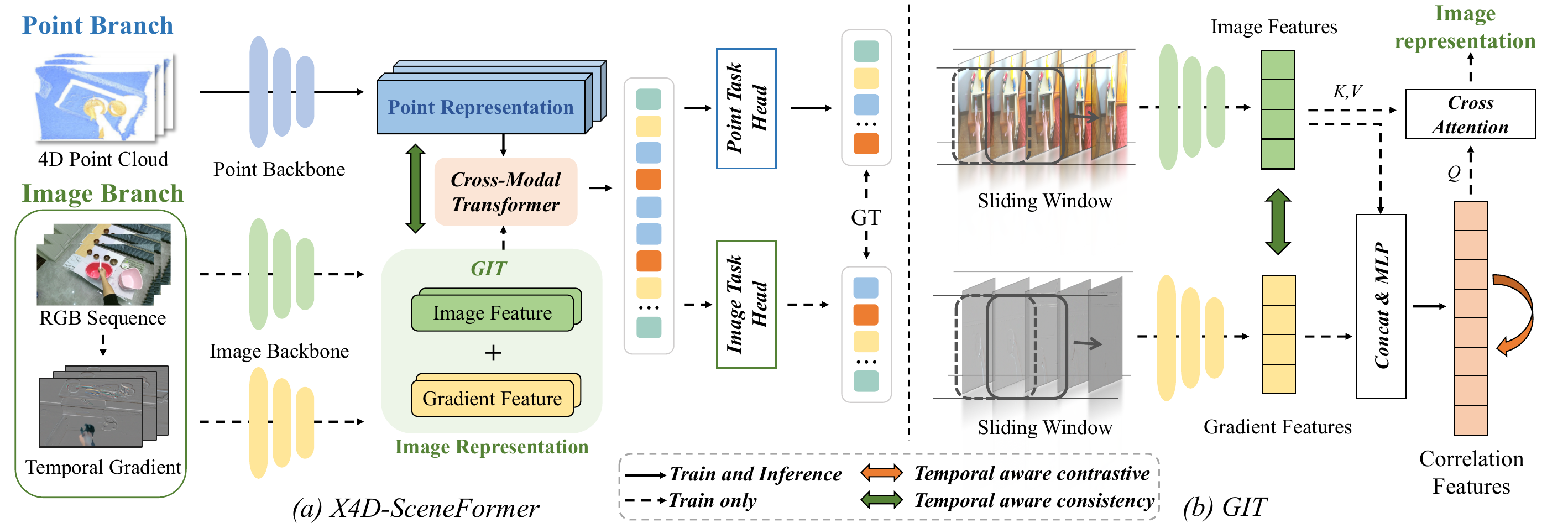}
\end{center}
\caption{\textbf{The architecture of \modelname, and GIT.} (a) During the training phase, \modelname, takes both image sequence and 4D point cloud as input, where the dual branches independently extract representations and are supervised by ground truths. A cross-modal Transformer process is applied between two representations. (b) The Gradient-aware Image Transformer (GIT) employs a sliding window strategy to establish temporal relationships and acquires a correlation feature through the cross-attention. Moreover, GIT applies two temporal-aware criteria in its processes. }
\label{fig:architecture}
% %\vspace{-.3cm}
\end{figure*}

\section{Related Works}

\subsection{Image-based Video Analysis}
%
% Deep neural networks have achieved excellent results in RGB/RGBD video understanding.
%
Previous image-based video analysis approaches~\cite{crasto2019mars} extract the global feature via RNN or 1D CNN~\cite{lea2017temporal}. 
After that, the following works enhance the performance through using two-stream network~\cite{ju2023distilling},
pooling techniques~\cite{fernando2016rank} and extracting averaged features from stridden sampled frames~\cite{wang2016temporal}.
%
% Some approaches, such as learn temporal relations from offline estimated optical flow with a separate branch of the network besides the RGB to capture both appearance and motion information. 
% %
% Besides,  are also introduced to select and merge frames into the video representation.
% %
% Temporal Segment Networks (TSN) is used to extract averaged features from stridden sampled frames.
%
In contrast, 3D CNNs~\cite{fernando2016rank} or sparse 3D convolution~\cite{graham20183d} jointly learn spatial-temporal features by organizing 2D frames into 3D structures to learn temporal relations implicitly.
%
% Recently, Self-attention has been extensively employed in computer vision tasks~\cite{vaswani2017attention}.
%
Recently, Vision Transformer (ViT)~\cite{dosovitskiy2020image} proposes a pure transformer architecture replacing all convolutions with self-attention, and achieved excellent results. 
Built on the ViT architecture, Timesformer and ViViT~\cite{arnab2021vivit} extend 2D spatial self-attention to the 3D spatial-temporal volume.

\subsection{4D Point Cloud Processing}
% 4D Point cloud video modeling is a novel yet consequential task, bearing pivotal significance for imbuing intelligent agents with the capacity to comprehend the ever-evolving 3D environment that envelops us. 
%
There are two mainstreams for 4D Point cloud video modeling: (1) voxel-based and (2) points-based approaches. 
\textbf{Voxel-based methods} first convert 4D point cloud into 2D voxel sequences, subsequently leveraging 3D convolutions to extract sequential features. For instance, MinkowskiNet~\cite{choy20194d} harnesses 4D sparse convolution, effectively mining features from valid voxel grids. 
3DV~\cite{wang20203dv} employs temporal rank pooling to fuse point motion within voxel sets, thereafter employing PointNet++~\cite{qi2017pointnet++} to extrapolate point representations. 
On the other hand, traditional \textbf{points-based methods} take raw point cloud as input, and exploits RNN~\cite{fan2019pointrnn}, appending a temporal features~\cite{liu2019meteornet} and  point spatial-temporal convolutions~\cite{fan2022pstnet} to encode temporal features.
%
% For example, PointRNN~\cite{fan2019pointrnn} introduces a point recurrent neural network tailored for predicting the trajectory of moving point cloud sequences. MeteorNet~\cite{liu2019meteornet} takes raw point cloud sequences as input, appending a temporal dimension to PointNet++ to holistically process 4D points. PSTNet~\cite{fan2022pstnet} proposes point spatial-temporal convolutions, fostering comprehensive information representation within point cloud sequences. 
%
Nevertheless, the above methods only focus on static scene representations. 
Recently, P4Transformer~\cite{fan2021point} introduces 4D point coevolution and then learns the temporal features in a Transformer architecture.
Building on this, PPTr~\cite{wen2022point} further boosts the performance  by incorporating primitive planes as prior knowledge, thereby enhancing the capture of enduring spatial-temporal context in 4D point cloud videos. 
PST-Transformer~\cite{fan2022point} encodes spatio-temporal structure by utilizing video-level self-attention to search related points adaptively. 
Notwithstanding these advancements, the existing methods typically cater to sparse and texture-limited point cloud inputs, ignoring rich texture and motion information in 2D images.

% The perception of videos can be approached from various modalities, each modality encodes unique insights. Learning from multiple modalities can provide rich learning information that enable the extraction of semantic and motion information from a given point cloud video. Recent investigations have demonstrated the efficacy of cross-modal learning, amalgamating distinct modality information to achieve superior outcomes compared to relying solely on a single modality~\cite{}. Through the synergistic integration of modalities, cross-modal networks can achieve enhanced performance. However, this advantage is accompanied by heightened computational requirements and latency during inference.

\subsection{Cross-Modality Learning and Knowledge Transfer}
Given that point cloud and images are capable of capturing distinct and complementary information pertaining to a scene, significant endeavors~\cite{yan20222dpass, afham2022crosspoint} have been undertaken to integrate multi-modal features in order to enhance perception.
However, the integration of multi-modal methods inevitably introduces additional computational burden and requires additional network design. As a result, recent works have focused on developing stronger single-modal models through the cross-modal knowledge transfer. 
Typically, knowledge transfer (KD) was originally proposed to compress integrated classifiers (teacher) into smaller networks (student) without significant performance loss~\cite{hinton2015distilling}. 
%
% Over the years, the effectiveness of knowledge transfer has led to its widespread exploration in various computer vision tasks, such as object detection~\cite{chen2017learning} and semantic segmentation~\cite{liu2019structured}.
%
Recently, KD has been extended to 3D perception tasks for transferring knowledge across different modalities.
Several approaches have been proposed for 3D object detection~\cite{wang20203dv}, 3D semantic segmentation~\cite{hou2022point}, and other tasks~\cite{yang2021sat}.
%
% Among these distillation methods, the most common strategy is to conduct alignments between the features and outputs of teacher and student networks~\cite{chen2020learning}.
%
Moreover, there are some previous approaches utilize contrastive criterion~\cite{zhang2023complete} to enhance the knowledge transfer during the training phrase.
Inspired by these works, we first time investigate cross-modal knowledge transfer in the task of 4D point cloud analysis. In contrast with previous methods that solely focus on distilling static spatial information, our architecture integrate motion and temporal alignment during the knowledge transfer.

% \noindent\textbf{Contrastive learning.} 
% Contrastive learning methods have demonstrated remarkable efficacy in unsupervised 2D and 3D tasks~\cite{}. Recent investigations~\cite{} have also demonstrated that label information can enhance contrastive learning within the supervised context. The core principle underlying contrastive learning revolves around learning representations by minimizing the feature distance between positive pairs (transformations of the same data sample) while simultaneously maximizing the feature distance between negative pairs (distinct data samples). Building upon this foundation, some researchers have proposed temporal contrastive learning techniques for self-supervised video learning~\cite{}, an inspiration that has led us to incorporate contrastive loss into 4D tasks.

\section{Methods}
\label{sec:MD}
In this paper, we introduce a novel cross-modality knowledge transfer approach that employs texture and motion priors to assist 4D point cloud analysis. As shown in Figure~\ref{fig:architecture}, our architecture consists of two branches, where the upper branch takes the normal 4D point cloud analysis model as the backbone while the other exploits extra RGB sequence to extract the prior knowledge. After that, we utilize a cross-modal Transformer to transform the cross-modal knowledge with masked attention. Besides, several knowledge transfer constraints are applied between the two modalities.

\subsection{Problem Formulation}
\label{subsec:Definitions}
The task of 4D point cloud analysis takes a point cloud video consisting of $T$ frames with $N$ points  as input, which can be denoted as $\mathcal{P}\in \mathbb{R}^{T\times{N}\times 3}$.
Typically, there are three main tasks in the 4D point cloud analysis: 4D semantic segmentation, action segmentation and action recognition.
The description of the above tasks can be formulated as

%\vspace{-.3cm}
\begin{align} \label{eqn:problem}
	\text{SemSeg}&: \mathbb{R}^{T \times N \times 3} \mapsto  \mathbb{R}^{T \times N }, \\
	\text{ActionSeg}&: \mathbb{R}^{T \times N \times 3} \mapsto  \mathbb{R}^{T}, \\
		\text{ActionRecog}&: \mathbb{R}^{T \times N \times 3} \mapsto  \mathbb{R}^1, 
\end{align}
where the former two segmentation tasks perform classification on point and frame levels respectively, and the recognition task identify single action for the whole video.

To assist the single-modal model during the training stage, we introduced RGB sequence as an additional input, denoted as $\mathcal{I}\in \mathbb{R}^{T\times{H}\times{W}\times 3}$  with the size of $H\times{W}$.
Taking 4D semantic segmentation as an example, the above task will be modified during the training:

%\vspace{-.3cm}
\begin{align} \label{eqn:problem2}
	\text{SemSeg}&: \mathbb{R}^{T \times N \times 3} \times \mathbb{R}^{T\times{H}\times{W}\times 3} \mapsto  \mathbb{R}^{T \times N }.
\end{align}
During the inference, the 4D point cloud model can be independently deployed and the formulation keeps the same as Eqn.~\eqref{eqn:problem}.

%

% \begin{figure}[t] 
% \begin{center}
% \includegraphics[width=0.7\linewidth]{Figure/Figure3.pdf}
% \end{center}
% \caption{\textbf{Temporal-aware consistency loss.} It aligns temporal information between two representations in a time-misaligned manner.}
% \label{fig3}
% \end{figure}

\subsection{4D Point Cloud Architecture}
\label{subsec:4D-rep}
The architecture of the 4D point cloud model (Point Backbone) is illustrated in the left section of Figure~\ref{fig:architecture}. Following the previous works~\cite{fan2021point}, We adopt point 4D convolution (P4Conv) as the encoder, generating the 4D point features with the shape of $\mathcal{F}^{P}_{l} \in \mathbb{R}^{T \times M \times D}$, where $M$ and $D$ are a number of subsampled points and channels, respectively. 
After that, several self-attention layers are applied to extract the sequential information across the sequence dimension. 
The outcome is a D-dimensional high-level feature representation, denoted as $\mathcal{F}^{P}_{h} = \left\{f_1, \cdots, f_t\right\}_{t=1}^T$.

\subsection{Gradient-aware Image Transformer (GIT)}
\label{subsec:2D-rep}
As described in the right part of Figure~\ref{fig:architecture}, Gradient-aware Image Transformer (GIT) is proposed to extract texture and gradient-aware features from the RGB sequence. It takes a set of images as input, independently encodes texture and gradient features and finally generates a high-level image feature representation $\mathcal{F}^{I}_{h}$ with a cross-attention module. Temporal-aware consistency and contrastive learning are applied during the training to enhance performance. 

\noindent\textbf{Gradient-aware feature encoding.} 
Inspired by previous work~\cite{xiao2022learning} exploiting temporal gradient (TG) to encode the sequential features, we first generate TGs as an extra input for the GIT.
The formulation of TG can be depicted as $g_t=I_{t} -I_{t+n}$, where $t$ denotes the frame index, $n$ is a predefined interval number, and $I$ is a section of RGB video. 
Given the input RGB video $\mathcal{I}$ and generated temporal gradient $\mathcal{G} = \{g_t\}_{t=1}^T$, two encoders adopt the same 2D-CNN architecture to extract low-level frame-based features $\mathcal{F}^{I}, \mathcal{F}^{G} \in R^{T \times D}$. 
\begin{figure}[t] 
\begin{center}
\includegraphics[width=0.9\linewidth]{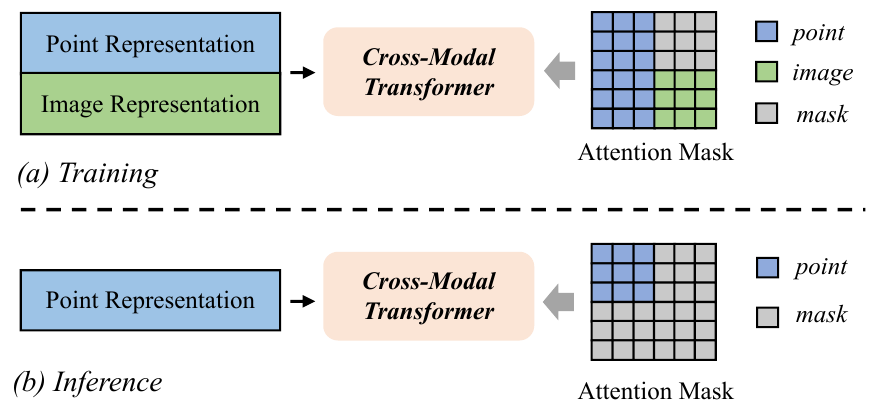}
\end{center}
% %\vspace{-.3cm}
\caption{\textbf{Masked attention in the cross-modal transformer.} The attention mask prevents point representation $\mathcal{F}^{P}_{h}$ from attending on image representation $\mathcal{F}^{I}_{h}$ in training (top three rows of the mask), avoiding performance drop in inference when $\mathcal{F}^{I}_{h}$ is not available.}
\label{fig4}
% %\vspace{-.3cm}
\end{figure}

\noindent\textbf{Fusion by sliding window.}
Since TG is a weak signal that cannot fully represent motion information, we further propose a sliding window mechanism to generate a fused correlation feature by mining the temporal relationship within $\mathcal{F}^{I}$ and $\mathcal{F}^{G}$.
Given the RGB and TG feature $\mathcal{F}^{I}=\{f_t^I\}_{t=1}^T$ and $\mathcal{F}^{G}=\{f_t^G\}_{t=1}^T$, the sliding window at the t-th time-step can be described as:

%\vspace{-.3cm}
\begin{align} \label{eqn:sw}
	\hat{f}_t^{I} &= \alpha_{t-n}*{f}_{t-n}^{I} + ... + \alpha{f}_{t}^{I} + ... +\alpha_{t+n}*{f}_{t+n}^{I},\\
	\hat{f}_t^{G} &= \beta_{t-n}*{f}_{t-1}^{G} + ... + \beta{f}_{t}^{G} + ... + \beta_{t+n}*{f}_{t+1}^{G},
\end{align}
where $\alpha$ and $\beta$ represent learnable parameters that assign weight to the motion trajectory at the boundary of actions. $n$ is the window size.
Subsequently, we merge the outputs of the sliding window and employ an MLP to generate a gradient-aware correlation feature $\mathcal{F}^{cor}$:
\begin{align} \label{eqn:fuse}
	\mathcal{F}^{cor} = \text{MLP}([\hat{\mathcal{F}}^{I};\hat{\mathcal{F}}^{G}]),
\end{align}
where $[\cdot;\cdot]$ is a concatenation operation.

\noindent\textbf{Temporal-aware contrastive.}
To improve the differentiation between various actions within a single sequence and address over-segmentation challenges in action segmentation tasks, we exploit a temporal-aware supervised contrastive loss on the aforementioned correlation feature $\mathcal{F}^{cor}$. 
Given a set of point cloud/label pairs with a temporal length of $T$ frames, denoted as $\left\{\mathcal{P}_i, \mathcal{Y}_i\right\}_{i=1,\ldots,T}$, a sequence of point cloud with various data augmentations can be represented as $\hat{\mathcal{P}}$, and the correlation features generated by $\hat{\mathcal{P}}$ are denoted as $\hat{\mathcal{F}}^{cor}$. 
Subsequently, we concatenate the aforementioned two predictions along the temporal dimension and denote it as $\boldsymbol{F}^{cor}$. 
The temporal-aware contrastive loss is formulated as follows:

%\vspace{-.3cm}
\begin{align}
	l(k, u) &= -\log \frac{\exp \left(\boldsymbol{F}^{cor}_k \cdot \boldsymbol{F}^{cor}_{u} / \tau\right)}{\sum_{{j} \in A(k)} \exp \left(\boldsymbol{F}^{cor}_k \cdot \boldsymbol{F}^{cor}_{j} / \tau\right)},\\
	\mathcal{L}^{{tcont}}&=\sum_{k \in M} \frac{1}{|G(k)|} \sum_{{u} \in G(k)} l(k, u).
	\label{eq:loss_C}
\end{align}
Here, $M = [1, 2T]$ is defined by the length of the concatenated sequence and $A(k) = M\backslash\{k\}$. $G(k) = \left\{{u} \in A(k): \mathcal{Y}_{u}=\mathcal{Y}_k\right\}$ denotes the set of positive pair and $\tau$ is a coefficient temperature. 
By employing this approach, the aforementioned loss function not only guarantees the proximity of features belonging to the same category within a given sequence but also facilitates the convergence of features from the same frame that has been augmented using distinct data augmentation.

\begin{table*}[t]
\centering
\small
\caption{The performance of action segmentation on HOI4D validation set and benchmark (CVPR2023-W) challenge. ${^1}$1st solution on HOI4D challenge. ${^2}$Runner-up solution in the challenge. 
${^3}$ We achieve 3rd place without using GIT module. 
%
% Since the top honors are not published, there is no disclosure of the authors’ personal information.
}
\begin{tabular}{l|c|ccccc|ccccc}
\hline
\multirow{2}{*}{Method} & \multirow{2}{*}{Reference} &
  \multicolumn{5}{c|}{Test} &
  \multicolumn{5}{c}{Validation} \\
 &&
  Acc &
  Edit &
  \multicolumn{3}{c|}{F1@\{10, 25, 50\}} &
  Acc &
  Edit &
  \multicolumn{3}{c}{F1@\{10, 25, 50\}} \\ \hline
P4Transformer &{CVPR 2021} &
  71.2 &
  73,1 &
  73.8 &
  69.2 &
  58.2 &
  63.2 &
  65.4 &
  65.9 &
  59.9 &
  45.9 \\
% PPTr &{ECCV 2022} &
%   77.4 &
%   80.1 &
%   81.7 &
%   78.5 &
%   69.5 &
%   72.3 &
%   75.6 &
%   74.8 &
%   70.3 &
%   58.4 \\
PPTr+C2P &{CVPR 2023} &
  81.1 &
  84.0 &
  85.4 &
  82.5 &
  74.1 &
  - &
  - &
  - &
  - &
  - \\
Multi-Conv-Res$^{2}$ &{HOI4D}&
  84.3 &
  86.6 &
  88.9 &
  86.9 &
  80.7 &
  - &
  - &
  - &
  - &
  - \\
DPMix$^{1}$ &{HOI4D}&
  85.2 &
  87.8 &
  89.8 &
  88.3 &
  82.9 &
  - &
  - &
  - &
  - &
  - \\ \hline
PPTr (Baseline)&ECCV 2022&
  77.4 &
  80.1 &
  81.7 &
  78.5 &
  69.5 &
  72.3 &
  75.6 &
  74.8 &
  70.3 &
  58.4 \\
\textbf{\modelname,$^{3}$ } & HOI4D &
  84.1 &
  91.1 &
  92.5 &
  90.8 &
  84.8 &
  78.9 &
  89.4 &
  88.2 &
  85.1 &
  75.1 \\
\textbf{\modelname, } &AAAI 2024&
  \textbf{85.3} &
  \textbf{91.5} &
  \textbf{92.6} &
  \textbf{91.1} &
  \textbf{85.5} &
  \textbf{82.6} &
  \textbf{92.4} &
  \textbf{91.8} &
  \textbf{89.4} &
  \textbf{81.2} \\
\emph{Improvement} & -& {\color{blue}+7.9} & {\color{blue}+11.4} & {\color{blue}+10.9}  & {\color{blue}+12.6} & {\color{blue}+16.0}  & {\color{blue}+10.3}  & {\color{blue}+16.8}  & {\color{blue}+17.0}  & {\color{blue}+19.1}  & {\color{blue}+22.8}   \\\hline
\end{tabular}
\vspace{-.3cm}
\label{tab:Action}
\end{table*}

\noindent\textbf{Temporal-aware consistency.}
To effectively utilize different temporal cues within a single sequence, we draw inspiration from the concept of asymmetric contrastive learning\cite{zhang2023complete}. In this regard, we employ a temporal-aware consistency loss to align temporal information between $\mathcal{F}^{I}$ and $\mathcal{F}^{G}$. This further enhances the generated feature and facilitates the prediction of motion trajectory by capitalizing on the geometric consistency of adjacent frames.
%
% As shown in Figure~\ref{fig3}, 
Given the image and gradient feature $\mathcal{F}^{I}$ and $\mathcal{F}^{G}$, the temporal-aware consistency loss aligns the temporal feature in a time-misaligned manner, \textit{i.e.,} advance and lag. It can be described as follows: 

%\vspace{-.3cm}
\begin{align}
		\mathcal{L}^{{adv}}=-\sum_{i=2}^N \log \frac{\exp \left(f^{G}_{i-1} \cdot f^{I}_{i} / \tau\right)}{\sum_{i=2}^N \exp \left(f^{G}_{i-1} \cdot f^{I}_i / \tau\right)}, \label{eq:loss_C1}\\
		\mathcal{L}^{{lag}}=-\sum_{i=1}^{N-1} \log \frac{\exp \left(f^{G}_{i+1} \cdot f^{I}_{i} / \tau\right)}{\sum_{i=1}^{N-1} \exp \left(f^{G}_{i+1} \cdot f^{I}_i / \tau\right)}. \label{eq:loss_C2}
\end{align}

Finally, the temporal-aware consistency $\mathcal{L}^{tac}_{GI}$ is a linear combination between the above two losses: $\mathcal{L}^{tac}_{GI}= (\mathcal{L}^{{adv}} + \mathcal{L}^{{lag}}) / 2$.
In such a manner, we introduce a temporal consistency constraint between differen features.

\noindent\textbf{Gradient-aware feature generation.}
The GIT involves the utilization of cross-attention blocks to merge the original spatial image feature $\mathcal{F}^{I}$ with the correlation feature $\mathcal{F}^{cor}$, thereby incorporating both spatial and temporal features. 
Specifically, the query is generated from $\mathcal{F}^{cor}$, while the key and value are obtained from $\mathcal{F}^{I}$ during this process.
We describe the obtained high-level image representation as $\mathcal{F}^{I}_{h}$.
% \begin{equation}
% Z_t = \operatorname{Attn}_{\text {cross }}\left(\mathrm{Q}_t, \mathbf{K}_t, \mathbf{V}_t\right)
% \label{eq:sw}
% \end{equation}

\begin{figure}[t] 
\begin{center}
% \fbox{\rule{0pt}{2in} \rule{.9\linewidth}{0p
% \includegraphics[width=1\linewidth]{Figure/Fig_main_v5.pdf}
\includegraphics[width=\linewidth]{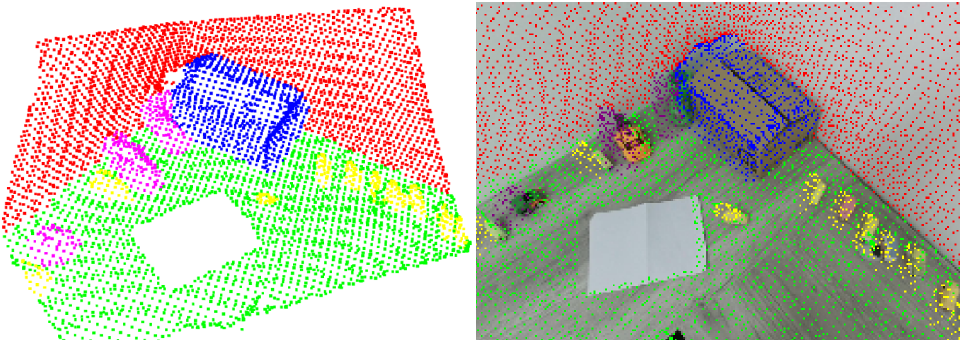}
\end{center}
% %\vspace{-.3cm}
\caption{\textbf{Visualization of GT generation for segmentation.} Since the 2D segmentation ground truths are not available in HOI4D dataset, we gain the the 2D labels through projecting the point cloud labels onto the image.}
\label{fig:Case1}
%\vspace{-.3cm}
\end{figure}

\subsection{Cross-modal Transformer}
\label{subsec:fusion}
To transfer the texture and gradient-aware knowledge from GIT to the 4D point cloud model, we design a cross-modal Transformer to fuse the knowledge from two modalities.
First, to ensure temporal consistency between the two modalities, we conduct temporal-aware consistency between $\mathcal{F}^{I}_{h}$ and $\mathcal{F}^{P}_{h}$, following a similar approach to Eqn.~\eqref{eq:loss_C1} and \eqref{eq:loss_C2}. This temporal consistency is denoted as $\mathcal{L}^{tac}_{PI}$.
We then employ a cross-modal transformer mechanism to merge their feature representations. 
Specifically, we adopt a stack of transformer layers to jointly encode the two input modalities $\mathcal{F}^{I}_{h}$ and $\mathcal{F}^{P}_{h}$. 
To avoid performance drop in inference when RGB sequence is not available, we design an attention mask inspired by \cite{yang2021sat}. As shown in Figure~\ref{fig4},  $\mathcal{F}^{P}_{h}$ does not directly attend to $\mathcal{F}^{I}_{h}$ (the top three rows of the mask). 
Meanwhile, the introduced attention mask allows the model to reference both $\mathcal{F}^{I}_{h}$ and $\mathcal{F}^{P}_{h}$ when generating the final output of the image branch (the bottom three rows of the mask). Finally, the output feature is utilized in several 4D task heads for downstream tasks, such as 4D action segmentation.

\noindent\textbf{Total loss functions.}
\label{subsec:Final_loss}
We denote $\mathcal{L}^{{P}}$ and $\mathcal{L}^{{I}}$ as the task supervision on the point cloud and image heads. The final loss can be described as:

%\vspace{-.3cm}
\begin{align}
\mathcal{L}= \mathcal{L}^{{P}} + \mathcal{L}^{{I}} + \omega * \mathcal{L}^{{tcont}} + 
(1-\omega) *  \mathcal{L}^{{tac}},
\label{eq:loss_final}
\end{align}
where $\omega$ denotes a hyper-parameter, and  $\mathcal{L}^{tac} =  \mathcal{L}^{tac}_{GI} +  \mathcal{L}^{tac}_{PI}$.

\begin{table}[t]
\centering
\footnotesize
\caption{4D semantic segmentation on HOI4D dataset.}
% We achieve \textbf{1st place} in HOI4D (CVPR2023-W) challenge.}
\begin{tabular}{l|c|cc}

\hline
\multirow{2}{*}{Method} & \multirow{2}{*}{Frames}&
  \multicolumn{1}{l}{\multirow{2}{*}{\begin{tabular}[c|]{@{}l@{}}~Test\\ mIoU\end{tabular}}} &
  \multirow{2}{*}{\begin{tabular}[c]{@{}c@{}}Val\\ mIoU\end{tabular}} \\
              &   & \multicolumn{1}{c}{}  \\
              \hline
P4Transformer  &3 & 40.1                  & 28.1 \\
PPTr+C2P            &10& 42.3                  & - \\\hline
PPTr (Baseline)          &3 & 41.4                  & 29.3 \\
\textbf{\modelname,}   &3& {\bf 47.3} & {\bf 35.8} \\ \hline
\end{tabular}

\label{tab:Semantic}
\end{table}

\section{Experiments}
\label{sec:exp}

\subsection{Experiments Setup}
\label{subsec:Setup}

{\noindent\bf Datasets.} 
We evaluate our proposed method on two benchmark datasets, namely HOI4D~\cite{liu2022hoi4d} and MSR-Action3D~\cite{li2010action}. The above datasets include three tasks: 4D action segmentation, 4D action recognition and 4D semantic segmentation.

The first dataset, HOI4D, contains 2,971 training videos and 892 test videos for action segmentation. Each video sequence has 150 frames with each frame containing 2048 points. The dataset contains a total of 579K frames. All frames are annotated with 19 fine-grained action classes in the interactive scene. Moreover, the 4D semantic segmentation task contains the same training and testing split with action segmentation. Each video sequence includes 300 frames of point clouds, with each frame consisting of 8192 points. Annotations involve 43 indoor semantic categories. The dataset contains a total of 1.2M frames.
Due to the non-public accessibility of the HOI4D test set, we randomly select 25\% of the training data as a validation set.

The second dataset, MSR-Action3D, consists of 567 human point cloud videos with 20 action categories. Each frame is sampled by 2,048 points. We maintain the same training/testing split as previous works~\cite{wen2022point,zhang2023complete}.

% Since the test set in HOI4D was no public, we randomly selected 25\% of the training data as the validation set and repeated it 10 times as the experimental result in both Action Segmentation and Semantic Segmentation tasks.

\begin{table}[t]
\centering
\footnotesize
\caption{Action recognition results on MSR-Action3D dataset. $^\star$We reproduce PPTr without using primitive fitting.}
\resizebox{\linewidth}{!}{
\begin{tabular}{l|c|c}
\hline
Method                          & Frames & Video Acc@1 \\ \hline
PointNet++~\cite{qi2017pointnet++}               & 1      & 61.61       \\ \hline
\multirow{3}{*}{MeteorNet~\cite{liu2019meteornet}}      & 8      & 81.14       \\
                                & 16     & 88.21       \\
                                & 24     & 88.50       \\ \hline
\multirow{3}{*}{PSTNet~\cite{fan2022pstnet}}         & 8      & 83.50       \\
                                & 16     & 89.90       \\
                                & 24     & 90.94       \\ \hline
\multirow{3}{*}{P4Transformer~\cite{fan2021point}}  & 8      & 83.17       \\
                                & 16     & 89.56       \\
                                 & 24     & 90.94       \\ \hline

\multirow{3}{*}{PPTr~\cite{wen2022point}}           & 8      & 84.02           \\
                                & 16     & 90.31          \\
                                & 24     & 92.33       \\ \hline                                
\multirow{3}{*}{PPTr+C2P~\cite{zhang2023complete}}           & 8      & \textbf{87.16}           \\
                                & 16     & 91.89           \\
                                & 24     & \textbf{94.76}       \\ \hline
\multirow{3}{*}{{PPTr$^\star$ (Baseline)}}           & 8      & 81.41           \\
                                & 16     & 90.87          \\
                                & 24     & 90.56       \\ \hline
\multirow{3}{*}{\textbf{\modelname,}}           & 8      &  86.47           \\
                                & 16     & \textbf{92.56}           \\
                                & 24     & 93.90       \\ \hline
\end{tabular}
}
\label{tab:a3Daction}
%\vspace{-.3cm}
\end{table}

{\noindent\bf Evaluation metrics.} 
For the task of  action segmentation, we exploit the metric of  frame-wise accuracy (Acc), segment edit distance (Edit), and segment F1 score with overlapping threshold k\% (F1@k) during the evaluation. 
Although frame-wise accuracy is commonly used as a metric for action segmentation, this measure is not sensitive to over-segmentation errors. 
The segmental edit score is presented in~\cite{lea2017temporal} and used to evaluate the case of over-segmentation, and the segmental F1 scores measure the quality of the prediction. 
For the task of 4D semantic segmentation, we rely on the mean Intersection over Union (mIoU) as our evaluation metric.
Finally, the top-1 accuracy is employed as the evaluation metric in the task of 3D action recognition.

{\noindent\bf Implementation details.} 
In our experiment, we employed ResNet18~\cite{jian2016deep} as the image encoders. For the point branch, we leveraged the point 4D convolution as the encoder, which was proposed by P4Transformer~\cite{fan2021point}. Across all experiments, the batch size was consistently set to 8. We used the SGD optimizer with a weight decay of $1\times 10^{-4}$. The learning rate was set to be 0.01 and we used a learning rate warm-up for 10 epochs, with a linear increase in the learning rate during this period. When calculating the contrastive loss, we set the temperature parameter $\tau$ to 0.07 and hyper-parameter $\omega$ to 0.5. The size of the sliding window was fixed at 3. 
In action segmentation task, the RGB sequence shares the same labels as the point cloud videos in each frame. Therefore, we follow \cite{fan2021point} and design linear layers as task heads for both modalities. 
As for the task of semantic segmentation, the task head are implemented with FP-layer~\cite{fan2021point} for point branch and de-convolution layers for image branch.
Since the 2D segmentation ground truths are not available in HOI4D dataset, we gain the the 2D labels through projecting the point cloud labels onto the image plane using point-to-pixel mapping. Figure~\ref{fig:Case1} illustrates an example of a point cloud scene along with its corresponding 2D projection. Subsequently, these projected 2D ground truths serve as the supervision for the image branch and only  labeled pixels are considered during the loss calculation.

\subsection{Comparison with State-of-the-arts}
\label{subsec:Compare_result}
%
% We evaluate our proposed framework against established methods on three challenging datasets: HOI4D Action Segmentation, HOI4D Semantic Segmentation, and MSR-Action3D. 
% %
%  For ensure result consistency for both Action Segmentation and Semantic Segmentation tasks, we repeated the experiment ten times. 

{\noindent\bf HOI4D action segmentation.} 
Table~\ref{tab:Action} demonstrates the results on HOI4D dataset for the task of action segmentation, where we compare our method with previously published methods~\cite{fan2021point,wen2022point,zhang2023complete} and other two unpublished methods on leaderboard (Multi-Conv-Res and DPMix).
\modelname, outperforms all comparative methods across evaluation metrics, on both test and validation sets. The test set results are sourced from the HOI4D online leaderboard. Its superiority is particularly evident in the metrics of edit distance and segment F1 score. Notably, P4Transformer and PPTr constitute the state-of-the-art backbones upon which other methods have further built. In particular, \modelname, exhibits improvements of at least 7.8\%, 11.4\%, and 10.9\% in terms of accuracy, edit distance, and F1@10 score respectively. The superior performance in edit and F1 scores demonstrates the effectiveness of our approach in over-segmentation issues, validating the effectiveness of our proposed temporal consistency metrics.

\begin{table}[t]
\centering
\footnotesize
\caption{Ablation study for different inputs. All experiments conducts without using GIT module.}
\begin{tabular}{ccc|ccccc}
\hline
\multicolumn{3}{c|}{Iputs} & \multirow{2}{*}{Acc} & \multirow{2}{*}{Edit} & \multicolumn{3}{c}{\multirow{2}{*}{F1@\{10,25,50\}}} \\
Point & RGB & TG &      &      & \multicolumn{3}{c}{} \\ \hline
$\checkmark$ &     &    & 72.3 & 75.6 & 74.8  & 70.3  & 58.4 \\
$\checkmark$  & $\checkmark$   &    & \textbf{77.5} & \textbf{76.4} & \textbf{75.7}  & \textbf{71.4}  & \textbf{59.5} \\
$\checkmark$  &     & $\checkmark$  & 72.8 & 76.1 & 75.2  & 70.6  & 58.9 \\
$\checkmark$ & $\checkmark$   & $\checkmark$  & 76.8 & 74.2 & 73.6  & 69.6  & 57.5 \\ \hline
\end{tabular}

\label{tab:abl_modal}
\end{table}

\begin{figure}[t] 
% \begin{center}
% \fbox{\rule{0pt}{2in} \rule{.9\linewidth}{0p
% \includegraphics[width=1\linewidth]{Figure/Fig_main_v5.pdf}
\includegraphics[width=0.9\linewidth]{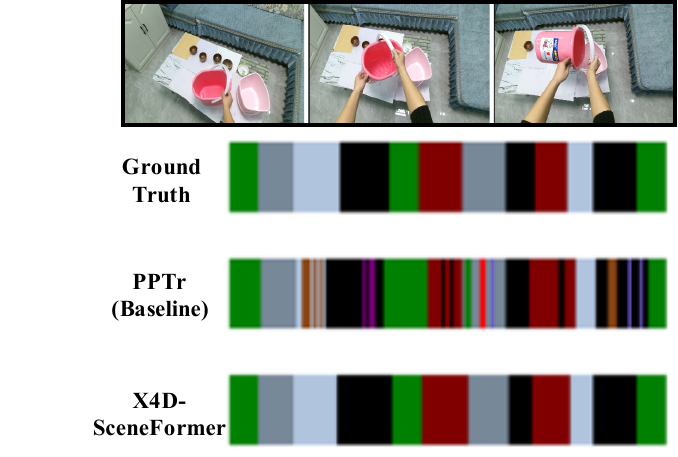}
% \end{center}
%\vspace{-.3cm}
\caption{\textbf{Visualization of action segmentation.} PPTr has a serious over-segmentation problem.}
%\vspace{-.3cm}
\label{fig:Case}

\end{figure}

{\noindent\bf HOI4D semantic segmentation.}
Table~\ref{tab:Semantic} provides the results, showing a mIoU of 47.3\% on the test set and 35.8\% on the validation set. The performance enhancement in the 4D semantic segmentation task highlights the efficacy of our approach in capturing fine-grained features. When compared to previous methods, our approach achieves superior results, which is attributed to the temporal alignment representation and robust generalization capabilities facilitated by cross-modal knowledge transfer and temporal consistency metrics.

{\noindent\bf MSR-Action3D.} 
The detailed results are presented in Table~\ref{tab:a3Daction}.
We reproduce the results of the PPTr without the primitive fitting as our baseline.
Considering the MSR-Action3D dataset lacks RGB data, we project the point clouds to the depth map as the input of the image branch.
Our approach demonstrates significant performance improvements across various sequence lengths. 
While our results are slightly below C2P~\cite{zhang2023complete}, this is primarily due to employing a weaker baseline and using projected depth as the input of image branch.
Still, we improve the performance upon baseline model by 5\%.
This observation underscores that \modelname, is not only well-suited to 4D tasks but also effectively addresses traditional video analysis.

\begin{table}[t]
\begin{center}
\footnotesize
\caption{Ablations study for GIT module.}
\scalebox{0.95}{
\begin{tabular}{c|c|ccccc}
\hline
    & Method                        & Acc  & Edit & \multicolumn{3}{c}{F1@\{10,25,50\}} \\ \hline
(a) & \modelname,-V         & 76.8 & 74.2 & 73.6       & 69.6       & 57.5      \\
(b) & + Correlation Feature         & 81.2 & 82.5 & 80.8       & 78.9       & 69.8      \\
(c) & + Sliding Window              & 81.9 & 84.5 & 82.3       & 81.1       & 72.6      \\ \hline
    % & \textit{Contrastive Learning} &      &      &            &            &           \\
(d) & + $\mathcal{L}^{{tcont}}$  & 82.2  & {87.9} & {86.4} & {85.2} & {76.3} \\
(e) & + $\mathcal{L}^{tac}$  & \textbf{82.6} & \textbf{92.4} & \textbf{91.8} & \textbf{89.4} & \textbf{81.2} \\ \hline
\end{tabular}

}
\label{tab:Ablation_GIT}
\end{center}
\end{table}

\begin{figure}[!t] 
\begin{center}
\includegraphics[width=0.8\linewidth]{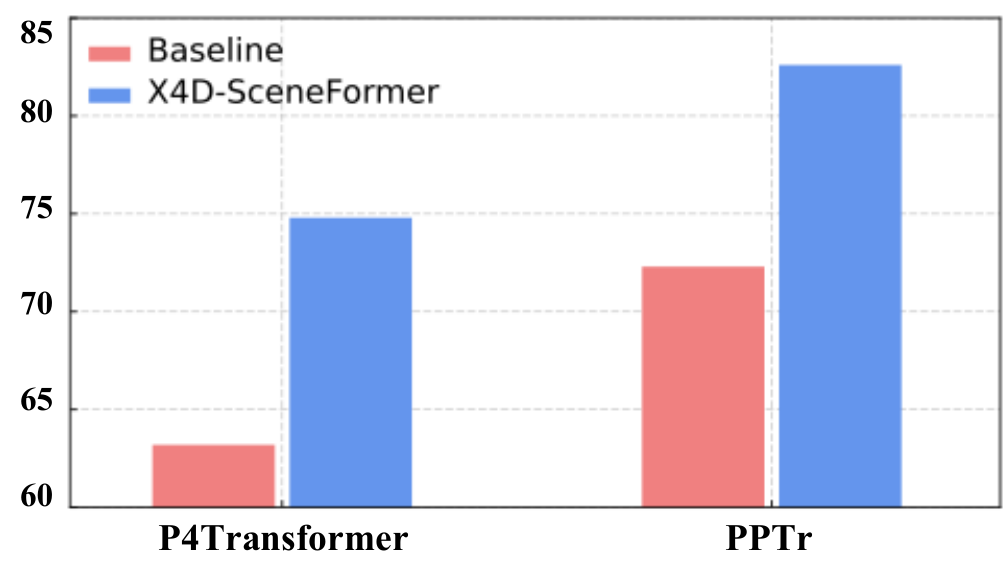}
\end{center}
\vspace{-.6cm}
\caption{Results using different 4D backbones.}
\label{fig:backbone}
\vspace{-.3cm}
\end{figure}

\subsection{Comprehensive Analysis}
\label{subsec:ablation}
All ablation experiments are conducted on HOI4D validation set with the task of action segmentation. 

{\noindent\bf The effect of different inputs.}
To validate the effectiveness of different input, we conducted ablation studies using various inputs \textbf{through replacing GIT module with simple concatenation}. 
As demonstrated in Table~\ref{tab:abl_modal}, using extra RGB sequence as input with the cross-modal transformer substantially improves performance, confirming the effectiveness of the cross-modal strategies.
However, naively increasing additional temporal gradient (TG) based on this foundation results in performance degradation.
We primarily attribute this to the temporal inconsistency between the two modalities. 
We name this model as \textbf{\modelname,-Vanilla}. Subsequently, we provide an explanation through follow-up experiments to illustrate the reasons behind the exceptional performance of our GIT method.

{\noindent\bf Design analysis of GIT.} Table~\ref{tab:Ablation_GIT} illustrates the effectiveness of each component in gradient-aware image Transformer (GIT). 
To improve the \modelname,-Vanilla (-V) discussed before , we generate the correlation feature through merging RGB sequence and TG with cross-attention. 
% This step becomes essential as the aforementioned experiments reveal that solely relying on TG is insufficient for effectively representing complex motion information.
%
The results demonstrate that the correlation feature significantly improves the performance, and the introduction of a sliding window further increase the result, especially for the edit distance with 2\% improvement. 
%
% Despite the correlation feature enhancing the performance of frame-wise accuracy (Acc), it still remains a challenge to improve the Edit distance and F1 score to more accurately represent the performance on long sequences.
%
Moreover, the introduced temporal consistency criterion lead to a substantial improvement in both edit distance (+8\%) and F1 scores (+9\%).
The outcomes demonstrate that the integration of cross-modal knowledge transfer and temporal consistency design effectively addresses the inherent over-segmentation challenge in 4D point cloud video tasks.

Table~\ref{tab:ablation_fusion} further illustrates the fusion strategy in GIT. 
It shows that cross-attention is the most effective manner of bridging RGB and TG features.
As illustrated in Figure~\ref{fig:Case}, our framework, incorporating GIT, demonstrates a superior capacity on HOI4D Action Segmentation dataset, especially for over-segmentation problem.

\begin{table}[t]
\footnotesize
\begin{center}
\caption{Ablation study for fusion mechanism in GIT.}
\begin{tabular}{l|ccccc}
\hline
Fusion         & Acc  & Edit & \multicolumn{3}{c}{F1@\{10,25,50\}} \\ \hline
concat         & 79.7 & 87.8 & 86.8       & 84.6       & 77.8      \\
sum.           & 79.5 & 87.6 & 86.5       & 84.3       & 77.5      \\
self-attention & 81.5 & 89.9 & 88.7       & 86.6       & 79.8      \\
\textbf{cross-attention} & \textbf{82.6} & \textbf{92.4} & \textbf{91.8} & \textbf{89.4} & \textbf{81.2} \\ \hline
\end{tabular}

\label{tab:ablation_fusion}
\end{center}
\vspace{-.3cm}
\end{table}

\begin{table}[t]
\footnotesize
\caption{Ablation study on various distillation baselines.}
\begin{tabular}{cc|ccccc}
\hline
    & Distillation      & Acc & Edit & \multicolumn{3}{c}{F1@\{10,25,50\}} \\ \hline
(a) & Transfer learning & 53.4   &   56.2   &     59.3       &    53.4        &      40.8    \\ 
    % & \textit{Teacher-Student}   &    &     &         &         &              \\
(b) &  L2 distance    &   61.2   &   61.1   &     63.5       &    58.2        &      45.6          \\
(c) & KL divergence       &  71.6   &   74.8   &    74.3        &    69.3        &      57.1     \\
(d) & Cosine similarity &  73.8   &   79.2   &    78.1        &    73.5        &       62.0    \\ \hline
% \multicolumn{1}{l}{(e)} & SAT           & \multicolumn{1}{l}{} & \multicolumn{1}{l}{} & \multicolumn{1}{l}{} & \multicolumn{1}{l}{} & \multicolumn{1}{l}{} \\
\multicolumn{1}{l}{(e)}                        & \textbf{Ours} & \textbf{82.6}        & \textbf{92.4}        & \textbf{91.8}        & \textbf{89.4}        & \textbf{81.2}        \\ \hline
\end{tabular}
% \vspace{-.3cm}
\label{tab:ablation_distill}
\end{table}

{\noindent\bf Different point backbones.} 
Figure~\ref{fig:backbone} demonstrates the results via using different point backbone. Our model respectively boosts the performance of P4Transformer and PPTr by 12\% and 10\%, which further verifies the generality of our proposed model.

{\noindent\bf Comparison of knowledge transfer.}
To further demonstrate the effectiveness of our cross-modal knowledge transfer framework, we conduct a series of experiments involving various classic distillation approaches.
As depicted in Table~\ref{tab:ablation_distill}, the application of transfer learning methods (a)~\cite{zhen2020deep} between the point branch and the image branch yields unsatisfactory results.
Furthermore, considering the widespread use of the teacher-student framework, we explored multiple experiments employing different distance functions between the modalities (b-d~\cite{hinton2015distilling}). However, despite the relatively improved performance of the cosine similarity loss, it still falls short of our proposed framework.
The primary factor is the temporal inconsistency inherent in two modalities. The results of all the ablation experiments consistently showcase the remarkable superiority of \modelname,.

%
% Additional ablation studies, encompassing changes in sliding window size, TG temporal frequency, 2D backbones, and visualization of semantic segmentation, are thoroughly explained in the supplementary material.

\section{Conclusion}
\label{sec:conclusion}
In this paper, we present \modelname,, a novel 4D cross-modal knowledge transfer framework that leverages texture priors from RGB sequences to enhance 4D point cloud analysis. 
Our framework consists of a 4D point cloud transformer and a Gradient-aware Image Transformer, which are trained with several knowledge transfer criteria to ensure temporal alignment and consistency between modalities. 
We show that our framework can achieve state-of-the-art results on various 4D point cloud video understanding tasks, such as action recognition and semantic segmentation, using only single-modal 3D point cloud inputs. 
Our work opens up new possibilities for 4D point cloud analysis that uses extra image priors to enhance performance.

\section{Acknowledgments}
This work was supported by the Shanghai AI Laboratory, National Key R\&D Program of China (2022ZD0160100) and the National Natural Science Foundation of China (62106183). Shenzhen General Program No. JCYJ20220530143600001, by the Basic Research Project No. HZQB-KCZYZ-2021067 of Hetao Shenzhen HK S$\&$T Cooperation Zone, by Shenzhen-Hong Kong Joint Funding No. SGDX20211123112401002, by NSFC with Grant No. 62293482, by Shenzhen Outstanding Talents Training Fund, by Guangdong Research Project No. 2017ZT07X152 and No. 2019CX01X104, by the Guangdong Provincial Key Laboratory of Future Networks of Intelligence (Grant No. 2022B1212010001), by the Guangdong Provincial Key Laboratory of Big Data Computing, The Chinese University of Hong Kong, Shenzhen, by the NSFC 61931024$\&$81922046, by the Shenzhen Key Laboratory of Big Data and Artificial Intelligence (Grant No. ZDSYS201707251409055), and the Key Area R$\&$D Program of Guangdong Province with grant No. 2018B030338001, by zelixir biotechnology company Fund, by Tencent Open Fund.

{
\bibliography{aaai23}
}

\newpage
\section{Supplementary}

% \section{Overview}
% \label{sec:evalution}
This document provides a list of supplemental materials to support the main paper. We illustrate more ablation studies on HOI4D action segmentation, the efficiency of the model, and the selected visualization cases. 

\section{Ablation Studies}
\label{sec:abl}

\noindent\textbf{Sliding window size.}
Our proposed sliding window strategy encourages the network not only to perceive motion information through geometric consistency but also to gain a comprehensive geometric understanding based on temporal motion cues.
Therefore, the window size requires careful consideration, as it determines the temporal knowledge incorporated into the correlation feature.
As detailed in Table~\ref{tab:window}, we conducted experiments on HOI4D action segmentation with sliding window sizes of 1, 3 (our setting), 5, and 7.
Setting the window size to 1 implies that both the RGB sequence and TG only consider the input from the current frame, disregarding information from preceding and subsequent frames.
The results, particularly the Edit and F1 scores, prove to be suboptimal due to the reliance solely on spatial information from a single time step. This can result in confusion between similar actions, leading to over-segmentation issues.
A sliding window size of 3 yields the best results, especially in Edit score (+7.9\%) and F1 scores (+8.6\%). This success can be attributed to its consideration of continuous actions over a specific time period.
Additionally, increasing the window size does not lead to substantial gains, and in some cases, it even results in lower performance. Larger time windows may span multiple actions, introducing redundant action information.

\begin{table}[b]
	\centering
	\caption{Ablation study for different sliding window sizes.}
	\label{tab:window}
	\begin{tabular}{c|ccccc}
		\hline
		Window\_Size & Acc & Edit & \multicolumn{3}{c}{F1@\{10,25,50\}} \\ \hline
		1            & 81.9 & 84.5 & 82.3       & 81.1       & 72.6       \\
		3            & \textbf{82.6} & \textbf{92.4} & \textbf{91.8} & \textbf{89.4} & \textbf{81.2}          \\
		5            & 81.5 & 84.1 & 81.7       & 80.6       & 71.4          \\
		7            & 79.8  & 82.1     &   78.5         &     77.3       &   68.8        \\ \hline
	\end{tabular}
\end{table}

\begin{figure}[t] 
	\centering
	\includegraphics[width=0.75\linewidth]{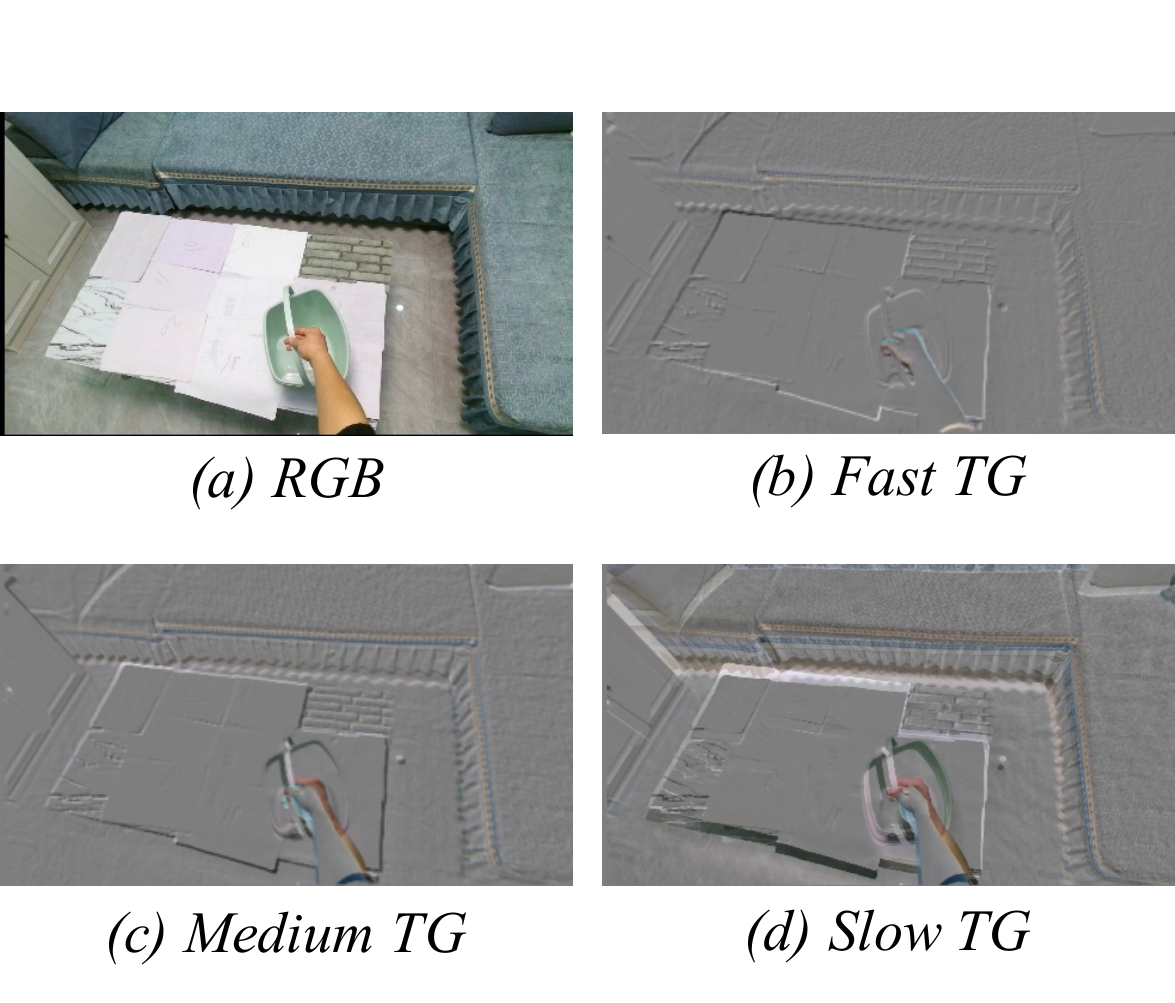}
	\caption{\textbf{Visualization of various TG rates.} The slow temporal gradient contains a more noisy background of the shooting environment, while the fast temporal gradient focuses more on the activity-related moving objects.}
	\label{fig:RGB_TG}
\end{figure}

\noindent\textbf{Temporal gradient (TG) rate.} 
Temporal gradient (TG) is computed by taking the difference between two RGB frames, and its stride can be either small or large, resulting in either fast or slow TG calculation. 
To gain deeper insights into the impact of different strides, we conducted ablation studies using fast TG (calculation stride = 1), medium TG (calculation stride = 4) and slow TG (calculation stride = 7).  
we visualized the three types of temporal gradient for HOI4D and the visualization are shown in Figure~\ref{fig:RGB_TG}.
The results, as presented in Table~\ref{tab:TG_n}, highlight a significant performance advantage of fast TG over other rates.

To figure out the reason why the performance of fast TG is much higher than slower TG, we visualized the three types of temporal gradient for the HOI4D dataset and the visualization is shown in Figure~\ref{fig:RGB_TG}.
The visualizations reveal that slower TG contains significantly more noisy background information, especially when cameras experience substantial movement. 
On the other hand, fast temporal gradient information primarily focuses on the boundaries of fast-moving objects.

\begin{table}[t]
\centering
\footnotesize
\caption{Ablation study for various TG rates.}
\label{tab:TG_n}
\begin{tabular}{c|ccccc}
\hline
Stride & Acc & Edit & \multicolumn{3}{c}{F1@\{10,25,50\}} \\ \hline
Slow              &  80.4 & 86.6 & 83.5       & 82.6      & 72.6       \\
Medium              &  80.2 & 82.3 & 82.3       & 81.1       & 72.6           \\
Fast             &  \textbf{82.6} & \textbf{92.4} & \textbf{91.8} & \textbf{89.4} & \textbf{81.2}          \\ \hline
\end{tabular}
\end{table}

% \usepackage{multirow}
\begin{table*}[t]
	\centering
\caption{Ablation study of the efficiency and effectiveness. The input `P' and `I' represent point cloud and image, respectively.}
\label{tab:train}
\begin{tabular}{c|cc|cc|c|cccc}
\hline
Method &Train Modal & Test Modal & Attention & Mask & Acc &\#Param (M, train/test)  & FPS (train/test) & FLOPS (M) \\ \hline
(a) &  P &   P   &          &  & 72.3 & 307.2 / 307.2   & 0.31 / 1.81 & 39732.1 \\
(b) &  P &    P  &       $\checkmark$   &  & 72.4 & 409.6 / 409.6 &  0.24 / 1.67 & 52174.6 \\
(c) & P+I  &   P    &   $\checkmark$       &  & 62.4 & 466.8 / 409.6 &  0.20 / 1.67 & 52174.6 \\
\textbf{(d)} & P+I  &   P    &    $\checkmark$      & $\checkmark$  & 82.6 & 466.8 / 409.6  & \textbf{0.20} / \textbf{1.67} & 52174.6 \\
(e) &  P+I &    P+I   &     $\checkmark$      & $\checkmark$ & \textbf{83.1} & 466.8 / 466.8  & 0.20 / 1.33 & 68241.6 \\ \hline
\end{tabular}
\end{table*}

\begin{figure*}[t] 
	\centering
	\includegraphics[width=0.85\linewidth]{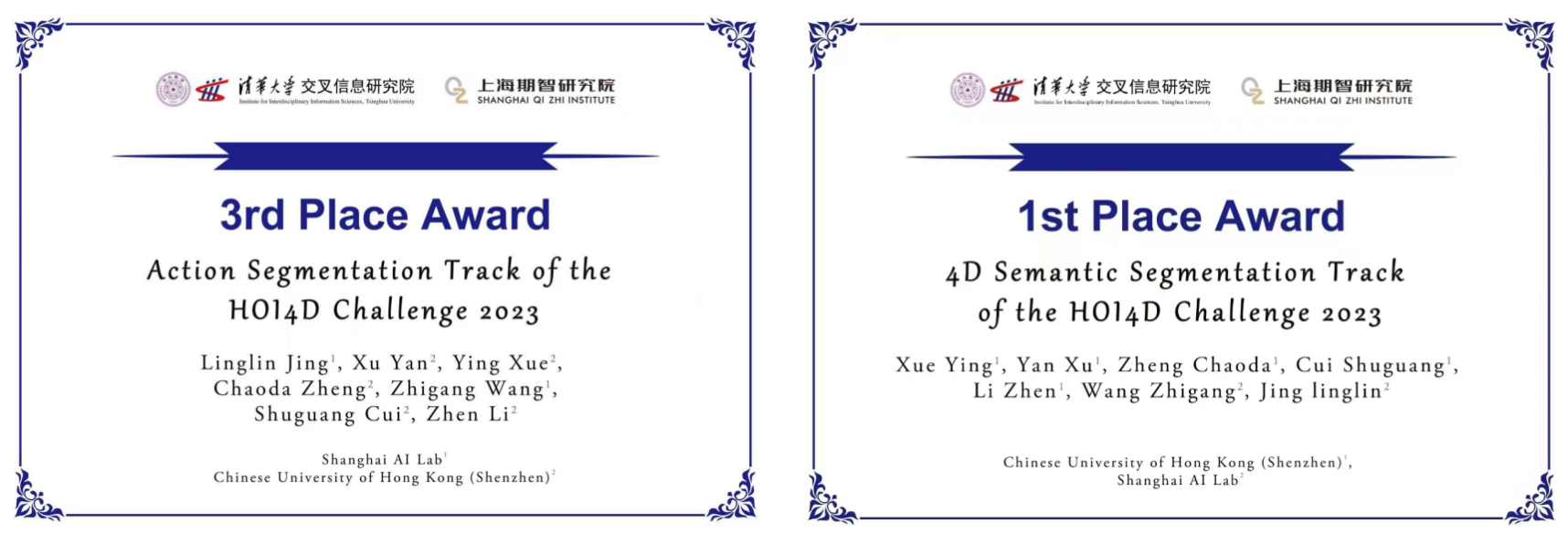}
	\caption{\textbf{Certificate of HOI4D competition.} The names and institutions have been obscured to ensure privacy.}
	\label{fig:Commendation}
\end{figure*}

\begin{figure}[t] 
	\centering
	\includegraphics[width=\linewidth]{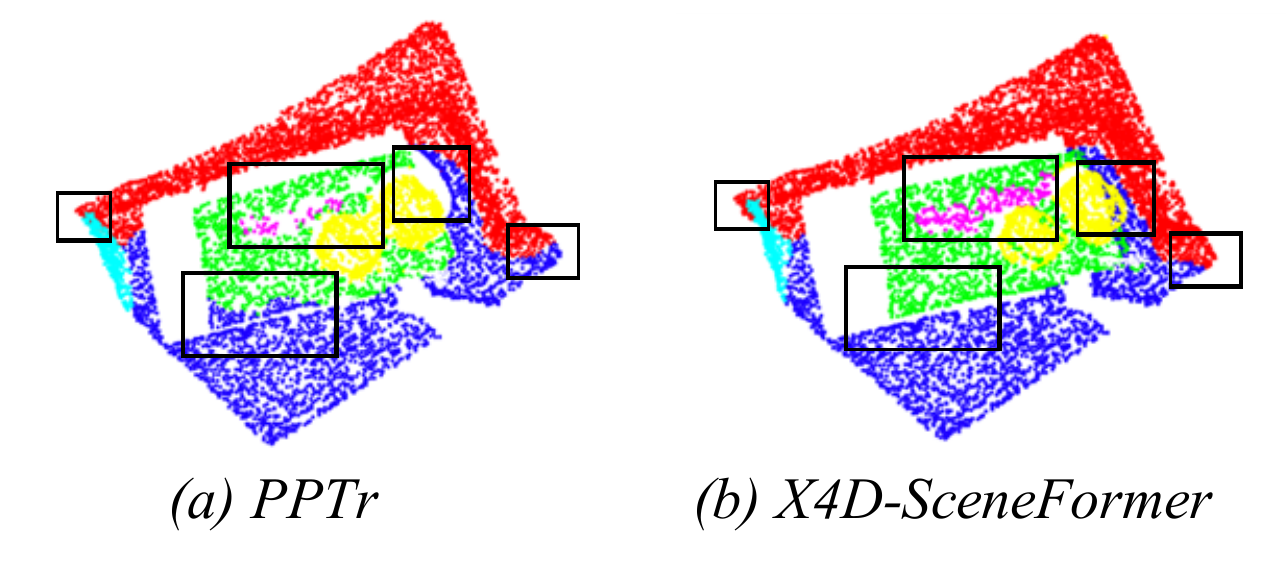}
	\caption{\textbf{Visualization of 4D semantic segmentation.} \modelname, outperforms PPTr (baseline), particularly in the segmentation of intricate scenes.}
	\label{fig:Semtic}
\end{figure}

\section{Effectiveness and Efficiency}
\label{sec:evalution}
We substantiated the effectiveness of knowledge transfer and the efficiency of the complete framework both during training and inference. Elaborated information can be found in Table~\ref{tab:train}. 
We present results from various experiments:

\noindent\textbf{(a)} The PPTr baseline with single point cloud input during both training and inference.

\noindent\textbf{(b)} The foundation of PPTr with self-attention layers, reflecting the network structure of \modelname, but ignoring additional image input.

\noindent\textbf{(c)} Introduction of image sequence during training, removing the attention mask to prevent RGB knowledge transfer to the point branch through a cross-modal transformer.

\noindent\textbf{(d)} \textbf{Our full model}, incorporating an attention mask in training to enable knowledge transfer between point and image.

\noindent\textbf{(e)} Image input used during both training and inference, resulting in the network structure being equivalent to that of training.

Notably, the simple addition of self-attention layers to the baseline (\textit{i.e.,} model (b)) doesn't yield performance improvements, underscoring that the success of \modelname, doesn't attribute to an increase in the number of parameters.
Furthermore, the absence of input data when an attention mask isn't used during training significantly reduces performance, as evidenced by (c).
In (e), the inclusion of image input improves performance. While (e) slightly outperforms (d), it introduces +15\% parameters and +30\% computational overhead during inference.
In summary, the design of \modelname, facilitates knowledge transfer between image and point during training, requiring only point input during inference. This design significantly enhances performance without introducing additional running time and computational resources during inference.

\section{Selected Visualization Case}
\label{sec:vis}
We give a visualization of HOI4D semantic Segmentation in this section to intuitively demonstrate the outstanding performance of our method. Illustrated in Figure~\ref{fig:Semtic}, within the intricate indoor scene, \modelname, exhibits superior performance with respect to intricate semantic distinctions (such as the green table, pink and yellow objects), resulting in an impressive mIoU score of 47.0\%. This achievement signifies a notable 5\% enhancement over the baseline performance.

\noindent\ Lastly, we have included the award certificate from HOI4D (CVPR-W 2023) to substantiate our accomplishment. For confidentiality reasons, the names and institutions have been obscured to ensure the privacy of the author's information.

\end{document}